\documentclass[11pt]{article}

\usepackage[preprint]{acl}

\usepackage{times}
\usepackage{latexsym}
\usepackage[T1]{fontenc}
\usepackage[utf8]{inputenc}
\usepackage{microtype}
\usepackage{inconsolata}
\usepackage{graphicx}

\usepackage{booktabs}
\usepackage{amsmath}
\usepackage{amssymb}
\usepackage{multirow}
\usepackage{xcolor}
\usepackage{float}
\usepackage{caption}
\usepackage{tabularray}
\usepackage{placeins}
\usepackage{enumitem}
\usepackage[most]{tcolorbox}

\newcommand{\code}[1]{{\small\texttt{#1}}}
\definecolor{goodgreen}{RGB}{0,180,0}
\definecolor{badred}{RGB}{220,30,30}

\definecolor{promptbg}{RGB}{245,245,250}
\definecolor{promptframe}{RGB}{180,180,200}
\definecolor{prompttitle}{RGB}{60,60,90}

\newtcolorbox{promptbox}[1][]{
  colback=promptbg,
  colframe=promptframe,
  coltitle=white,
  colbacktitle=prompttitle,
  fonttitle=\bfseries\small,
  fontupper=\ttfamily\scriptsize,
  boxrule=0.5pt,
  arc=2pt,
  left=4pt, right=4pt, top=2pt, bottom=2pt,
  title=#1
}

\newif\ifdraftmode\draftmodetrue
\ifdraftmode
  \providecommand{\psj}[1]{{\protect\color{red}{[Psj: #1]}}}
  \providecommand{\nishi}[1]{{\protect\color{blue}{[Nishi: #1]}}}
  \providecommand{\todo}[1]{{\protect\color{orange}{[TODO: #1]}}}
\else
  \providecommand{\psj}[1]{}
  \providecommand{\nishi}[1]{}
  \providecommand{\todo}[1]{}
\fi

\title{Translate-R1: Cost-Aware Translation Tool Use via Reinforcement Learning}

\author{
  Pratik Jayarao, Chaitanya Dwivedi, Himanshu Gupta, Neeraj Varshney, \\
  \textbf{Adithya M Devraj, Meet Vadera, Priyanka Nigam, Bing Yin} \\
  Amazon Stores Foundation AI \\
}

\begin{document}
\maketitle

\begin{abstract}

The performance gap across languages in LLMs is well documented, and closing it natively requires pretraining or fine-tuning on corpora that, for most languages, are quite limited. Translation offers an alternative: converting an input into the model's dominant language unlocks its full capabilities at once. But translating every input is wasteful for languages the model already handles, while leaving the choice to the model fails too, as LLMs are overconfident and skip the tool even when they cannot understand the input \citep{wang2025xwebagent}. Prior work resolves this with language-specific rules, heuristics, or external routers, each requiring manual engineering. We instead learn a single policy that decides when to translate from reward alone, developing language- and domain-adaptive introspection that invokes translation only when it cannot solve a task natively.

Using data from our answer-preserving translation pipeline, we continue RL on the post-trained Qwen3-4B across 22 languages in 3 resource tiers (High, Low, XLow) and 5 domains, and introduce confidence-gated GSPO for cost-sensitive tool use. The gated policy lifts reward over the untrained baseline by +4.6 on High, +23.5 on Low, and +17.5 on XLow, and extends zero-shot to 9 held-out languages. Against an unconstrained policy that almost always translates (the reward upper bound), it preserves full reward at \textbf{66\% of the cost}, outperforming other cost penalties by up to +24.3 on the low-resource tiers while matching the free policy's reward on high-resource tiers at a fraction of the tool use. On 2 synthetic languages with zero prior exposure it correctly learns to always translate, improving +18.7 over the overconfident baseline.
\end{abstract}

\begin{figure*}[t]
\centering
\includegraphics[width=\textwidth]{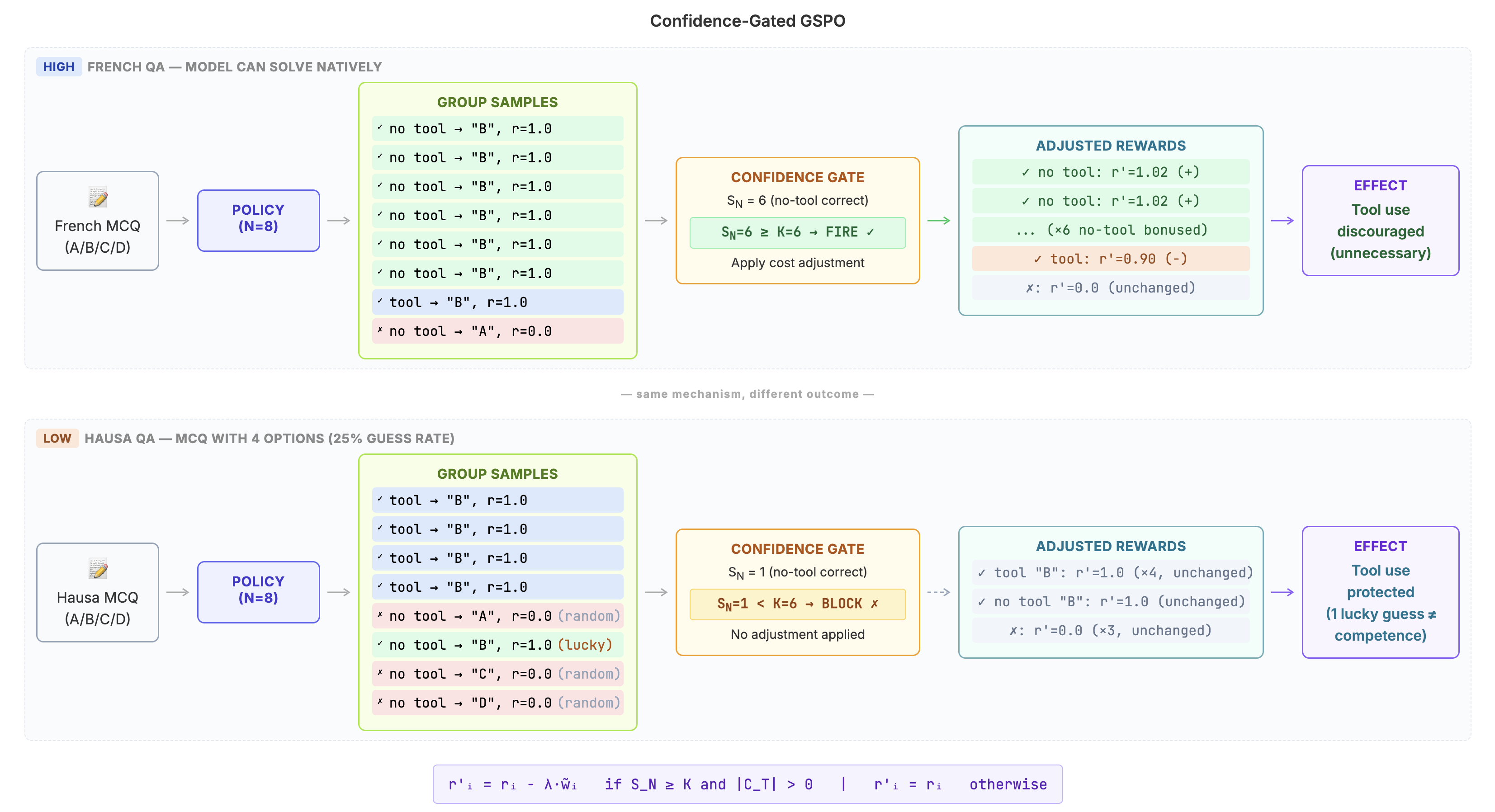}
\caption{The confidence-gated mechanism: the same group-relative rule penalizes unnecessary translation for high-resource (top) but protects necessary translation for low-resource (bottom), adapting via binomial statistics without language labels.}
\label{fig:gated_teaser}
\end{figure*}

\section{Introduction}

LLMs perform far better in some languages than others \citep{shi2023language, son2025multilingual, globalmmlu2024}, and meaningfully improving a weak language usually means pretraining or large scale fine-tuning on more of its data. Translation offers a different path: unlike tools that augment a single capability (search for knowledge, code for computation), it converts an incomprehensible input into one the model already reasons well in, unlocking its \textit{broader} capabilities at once.

However, translation brings its own dilemma. On one hand, applying it to every input is wasteful for languages the model already handles and can even hurt when the translator introduces errors. On the other hand, leaving the choice to the model does not work either, since models tend not to use the tool even when they should. This phenomenon has been studied by \citet{wang2025xwebagent}, who find that LLMs never invoke translation tools across 14 languages, forfeiting large gains. In most cases, whether translation helps is not fixed: it depends on the language, the domain, and the specific input, a per-sample decision that cannot be specified by hand.

Prior work makes this choice with language-specific rules, domain heuristics, language identifiers, or external routers \citep{routellm2024, xrouter2025}, all requiring manual engineering or auxiliary components. We ask whether the choice can instead emerge from reward alone. We learn a single policy across languages spanning high-resource to extremely scarce, across diverse domains, and under varying cost budgets. Trained only on task reward, it develops language- and domain-adaptive introspection, sensing its own competence and translating only when it pays off, with no language identification, routing rule, or supervised demonstration.

Our contributions:
\begin{enumerate}
    \item \textbf{A learned introspective policy.} From task reward alone, the model learns tool use that adapts to both language and domain: it translates math and QA in low-resource languages but solves instruction following natively in those same languages. Both behaviors emerge without explicit engineering.
    \item \textbf{Cost-sensitive tool use via Confidence-Gated GSPO.} Applying a fixed cost penalty cannot adapt to language difficulty: it over-suppresses tool use for exactly the low-resource languages that need it most, dropping their performance. We introduce a confidence gate that applies cost pressure only when the model demonstrates strong native competence, adapting automatically to language difficulty without tier labels. The resulting policy retains full unconstrained reward at 66\% of the cost, outperforming flat and OTC cost penalties by up to +24.3 on the low-resource tiers while matching the free policy's reward on high-resource tiers at a fraction of the tool use.
    \item \textbf{An answer-preserving translation pipeline.} A common way to build multilingual training data is to translate existing English datasets, but for RLVR this breaks a key invariant: the ground-truth answer must stay valid after translation. Naive translation of the RLVR pair $(q, a)$ can lead to corruption or inconsistency, and LLM judges cannot reliably verify in low-resource languages. Our pipeline sidesteps this by verifying entirely in the model's dominant language via back-translation, reaching 98.4\% fidelity (independently audited with Claude Opus) across our low-resource languages.
    \item \textbf{Tool use on completely unseen languages.} We also study how the model behaves on languages it has never seen, where the correct behavior is generally to translate, yet the overconfident baseline underutilizes the tool. To study this cleanly, we construct 2 synthetic languages with no possibility of prior exposure, where partial understanding is impossible. Our gated policy learns to recognize that it cannot comprehend the input and translates, gaining +18.7 points over the baseline.
\end{enumerate}

\section{Related Work}

\paragraph{Multilingual reasoning and RL.} The cross-lingual performance gap is well established \citep{shi2023language, son2025multilingual}, with the bottleneck often traced to comprehension rather than reasoning \citep{qalign2025, englishpivot2025}. One line of work bridges the gap by moving reasoning into English: cross-lingual prompting \citep{qin2023cross} and translate-then-solve training \citep{mathoctopus2024} do so ahead of time, while TAPO \citep{tapo2025} learns to pivot through English with reinforcement learning. A parallel line keeps reasoning in the source language, since RL with verifiable rewards, using GRPO \citep{deepseekmath2024} and its variants \citep{deepseekr1, dapo2025}, transfers across languages better than supervised fine-tuning \citep{beyondenglish2025} but can drift toward the dominant language \citep{crosscollapse2025}, motivating language-consistency rewards \citep{thinknatively2025}. Whether they translate or reason natively, these methods fix the strategy in advance or shape it globally through reward; none decide per input from the model's own competence. We instead learn a single policy that decides when to translate end-to-end from reward, jointly across languages and domains. Concurrent to our work, \citet{kang2026luar} study RL-based translation tool use for math reasoning.

\paragraph{Tool-use RL and cost-aware routing.} Models can be trained to call search or code tools via RL \citep{jin2025searchr1, li2025torl, feng2025retool, tora2024, steptool2024}, to decide when to invoke them and to call fewer of them \citep{tocodeornot2025, wang2025otc}, or to route between models and modes under a cost budget \citep{routellm2024, xrouter2025, routerr1}. In the multilingual setting, however, agents often leave translation tools unused even when available \citep{wang2025xwebagent} and degrade with resource level \citep{maps2025, massiveagents2025}. Translation also differs from search or code: the model may not understand its input at all, making the call a metacognitive judgment, and tool utility is language-conditional (useful in Hausa, less so in French). We adapt the group-relative optimization of Nemotron Nano \citep{nemotron2025} and Dr.\ GRPO \citep{drgrpo2025} from response length to tool cost, and add a confidence gate that prevents over-suppression on the low-resource languages that most need the tool.

\section{Data}
\label{sec:pipeline}

\subsection{Languages and Domains}

We select 22 natural languages across 3 resource tiers (High, Low, XLow), set by their approximate number of digital speakers, a proxy for representation in web-scale pretraining data. This proxy is coarse: it does not always track downstream performance, and an XLow language can outperform a Low one on a given task when it shares script, vocabulary, or close neighbors with high-resource languages (see Limitations). We also build 2 synthetic languages, \textit{Kivari} and \textit{Toqal}, to simulate zero pretraining exposure (Section~\ref{sec:cipher}). We train across 5 domains: 3 verifiable (math, QA, instruction following) with deterministic rewards, and 2 non-verifiable (summarization, translation) scored by an LLM judge. The full tier-to-language assignment is given in Table~\ref{tab:languages} (Appendix~\ref{app:data_sources}).

\subsection{Scalable Translation Pipeline for Verifiable Domains}
\begin{figure*}[t]
\centering
\includegraphics[width=\textwidth]{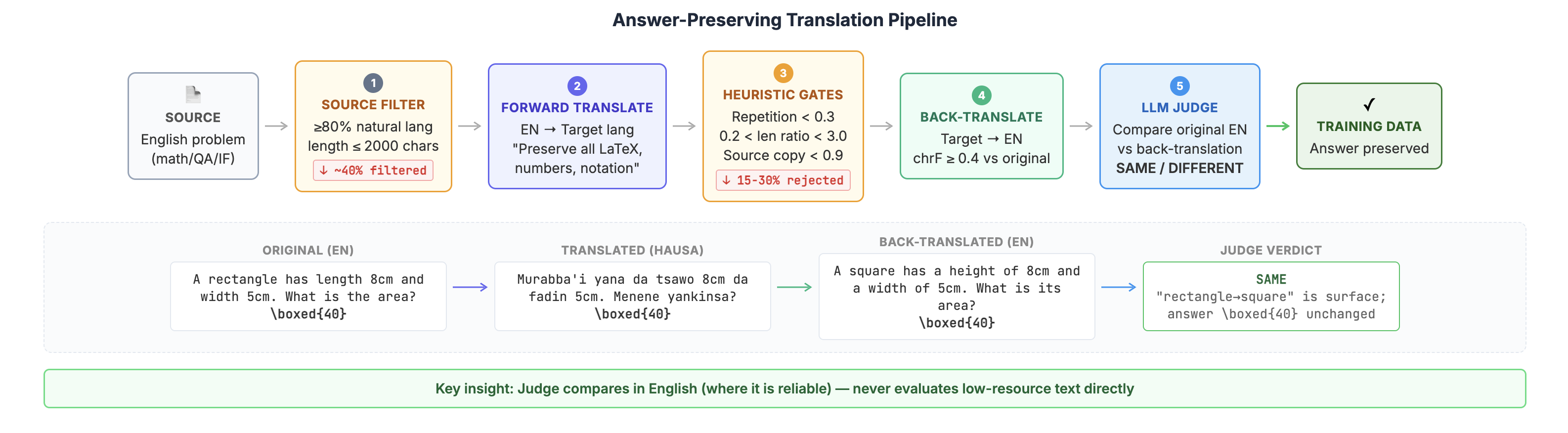}
\caption{The answer-preserving translation pipeline: verification runs entirely in English via back-translation, where the judge is reliable, rather than evaluating low-resource text directly.}
\label{fig:pipeline}
\end{figure*}

RLVR has become the backbone of modern reasoning post-training, and the natural way to extend it to more languages is to translate existing English datasets. But RLVR depends on an invariant that translation threatens: the ground-truth answer must stay valid after translation. Naive verification fails for low-resource languages, as surface metrics (BLEU/chrF) miss semantic corruption, direct LLM judges lack target-language competence, and solve-based checks are too expensive at scale. Our insight is to \textit{back-translate} to English and ask the judge to \textit{compare} rather than \textit{solve}, working in its strongest language. The pipeline runs five stages (Figure~\ref{fig:pipeline}), and every training sample passes all five:

\begin{itemize}[leftmargin=*, itemsep=1pt, topsep=2pt]
  \item \textbf{Source Filtering.} Drop problems with little natural-language content or excessive length, where translation is most error-prone.
  \item \textbf{Forward Translation.} Domain-aware prompts keep verbatim what must not change (notation, JSON structure, option keys) and translate only the natural-language parts.
  \item \textbf{Heuristic Filtering.} Cheap checks catch degenerate translations before the judge call: repetition, length-ratio bounds, source-copy detection.
  \item \textbf{Back-Translation + chrF.} Render the translation back to English and measure chrF against the original, capturing round-trip information loss.
  \item \textbf{LLM Judge.} Compare the original English with the back-translation for a binary SAME/DIFFERENT verdict, sidestepping unreliable low-resource judging (Appendix~\ref{app:pipeline_details}).
\end{itemize}

\subsection{Non-Verifiable Domains}

Translation data runs through the same five-stage pipeline with a meaning-preservation judge, whilst summarization uses native-language XL-Sum \citep{xlsum2021} articles where available (Appendix~\ref{app:nonverifiable}).

\subsection{Synthetic Languages: A Completely Unseen Language}
\label{sec:cipher}
To simulate a completely unseen language, where partial understanding is impossible and the only route to a correct answer is the translation tool, we construct two synthetic languages, \textbf{Kivari} and \textbf{Toqal}, by deterministic word-level substitution of English (construction details in Appendix~\ref{app:synthetic}). For these languages the tool is a deterministic lookup server returning perfect translations behind the same \code{<tool\_call>} interface.

\subsection{Data Distribution}

Within each domain, low-resource languages get 4--8$\times$ more samples per language than high-resource ones. The full training set has 273K samples across all domains and tiers (Tables~\ref{tab:data_dist_v} and~\ref{tab:data_dist}, Appendix~\ref{app:data_sources}). Evaluation uses a disjoint held-out split from the same sources, up to 50 samples per language and domain (Table~\ref{tab:eval_samples}, Appendix~\ref{app:eval_data}). An independent audit of the full evaluation set, with Claude Opus as annotator, confirms \textbf{98.4\% fidelity} across all Low and XLow languages (Appendix~\ref{app:audit}).

\section{Training Setup}
\label{sec:training}

\subsection{Model and Continued RL}
\label{sec:continued_rl}
All our training is \textit{continued} RL on top of the Qwen3-4B \citep{qwen3} post-trained checkpoint, run in reasoning mode. This checkpoint already has strong reasoning and basic tool calling abilities. Our goal is not to instill these abilities but to teach the model \textit{when} to invoke translation. Our \textbf{Baseline} is this checkpoint evaluated zero-shot, and every RL variant (no-tool, free, gated, flat, OTC) is continued RL from it under identical data and hyperparameters.

\subsection{Setup}
Each rollout involves three distinct components: the \textbf{agent} being trained, the \textbf{translator} it can call as a tool, and the \textbf{judge} that scores its output.

\paragraph{Agent.} The Qwen3-4B model being trained, which for each input decides whether to answer natively or emit \code{<tool\_call>} to leverage translation.

\paragraph{Translator (Tool).} A single Qwen3.5-122B-A10B server backs the translation tool. When the policy emits a \code{<tool\_call>}, the call runs and its result is injected back as a \code{<tool\_response>}, with a deterministic lookup server standing in for synthetic languages behind the same interface. A 4B policy calling a 122B translator is not a realistic deployment ratio, but using an off-the-shelf strong model lets us sidestep training bespoke MT systems per language pair. We deliberately use a strong 122B model here as a proxy for that service to guarantee reliable translations, particularly for the low-resource languages where weaker translators are error-prone, so that what we measure is the policy's when-to-translate decision rather than translator noise (Appendix~\ref{app:translator}). In a real deployment, this large model would be replaced by a competent but far smaller translation model.

\paragraph{Judge (Reward).} Rewards are deterministic for the verifiable domains (exact match on \code{\textbackslash boxed\{\}} for math, letter match for QA, keyword/constraint satisfaction times an LLM fluency gate for IF) and use the same Qwen3.5-122B-A10B model as a structured LLM judge for the non-verifiable ones (summarization, translation; Appendix~\ref{app:judge_prompts}).

All domains are mixed within each rollout under inverse resource weighting, and system prompts never reveal the input language, so the model must assess its own comprehension (Appendix~\ref{app:prompts}).

\subsection{Algorithm}

For all our experiments we use Group Sequence Policy Optimization (GSPO; \citealp{gspo2025}), and train without a KL penalty. (Appendix~\ref{app:hyperparams}).

\paragraph{Group-Relative Tool Efficiency.} With vanilla GSPO, the model learns to call the translation tool on nearly every input, including high-resource languages it already handles, since translating rarely hurts reward. To discourage this unnecessary tool use, we add a lightweight reward post-processing step inspired by Nemotron Nano \citep{nemotron2025} and OTC \citep{wang2025otc}: within each group of $N$ samples for one prompt, reward the most efficient correct solution. Among correct samples only, those with fewer tool calls get a zero-sum bonus and those with more take a penalty:
\begin{equation}
    r_i' = r_i - \lambda \cdot \tilde{w}_i \quad \forall\, i \in \mathcal{C}
\end{equation}
where $\tilde{w}_i$ is the zero-mean centered cost weight within the correct set and $\lambda$ controls the strength of cost pressure.

\paragraph{The Cascade Problem.} This mechanism, however, cascades into over-suppression for low-resource languages. When the model cannot comprehend the input, an occasional low-confidence no-tool answer still lands as correct, and this triggers the penalty on genuinely necessary tool use, forming a feedback loop that collapses tool adoption. We analyze this in Section~\ref{sec:cascade_analysis}.

\paragraph{Confidence-Gated GSPO.} To resolve the cascade, we propose \emph{Confidence-Gated GSPO}, which applies the group-relative cost penalty only after the group shows strong evidence of native competence (Figure~\ref{fig:gated_teaser}):
\begin{equation}
    r_i' = \begin{cases}
        r_i - \lambda \cdot \tilde{w}_i & \text{if } S_N \geq K \text{ and } |\mathcal{C}_T| > 0 \\
        r_i & \text{otherwise}
    \end{cases}
\end{equation}
where $S_N = \sum_{j: c_j=0} \mathbf{1}[r_j > 0]$ counts no-tool correct samples in the group and $\mathcal{C}_T = \{i \in \mathcal{C} : c_i > 0\}$ is the subset of correct samples that used a tool. The gate fires only when $K$ of $N$ no-tool samples are correct, so a few low-confidence correct answers no longer trigger suppression. Its firing probability follows a binomial that adapts to language difficulty without tier labels. We study the gate threshold $K$ and cost strength $\lambda$ in Appendix~\ref{app:gate_math}.

\section{Experiments}

Our experiments build up in three stages, each adding one degree of freedom so we can isolate its effect: first no tool at all, then an unconstrained tool, and finally a tool under cost pressure. This separates what continued RL achieves on its own from what the translation tool adds, and what a cost signal changes about how the tool is used.

\subsection{No Tool (Data Validation)}

We begin without any tool access: to measure the model's native ceiling without external help, and to validate that our translated data provides a clean reward signal. The model trains across all 22 languages and 5 domains and must solve every problem natively (Table~\ref{tab:no_tool}; analyzed in Section~\ref{sec:no_tool_analysis}).

\subsection{Free Tool (Unconstrained Translation)}
\label{sec:free_tool}

We then add the translation tool with no cost penalty, asking whether the model can learn \textit{when} to use it from task reward alone. The tool is described in the system prompt and may be called up to twice per response or not at all; every correct answer earns full reward regardless of tool use. The model adopts the tool quickly and improves reward across tiers (Table~\ref{tab:cost_focused}), but converges to translating almost everything, including high-resource inputs it already handles, so the gains come with wasteful over-translation.

\subsection{Cost-Sensitive Tool Use}
\label{sec:cost_sensitive}
\begin{figure*}[t]
\centering
\includegraphics[width=\textwidth]{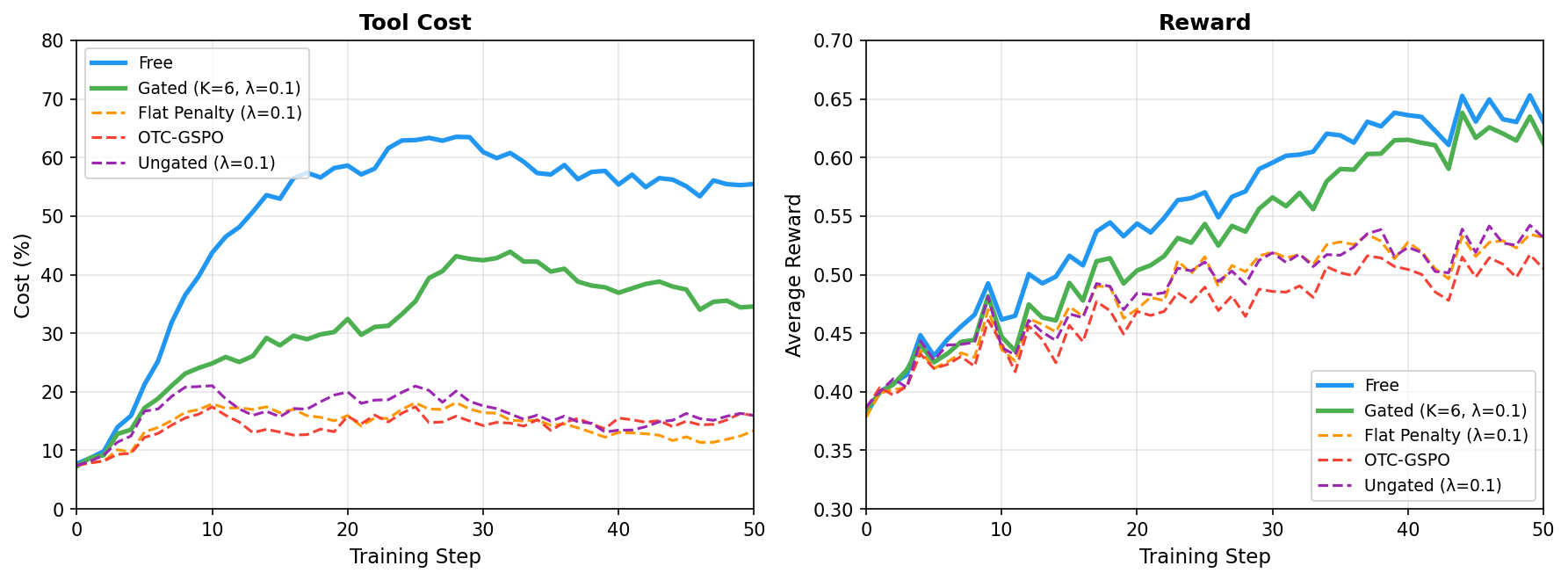}
\caption{Cost mechanism comparison (left: tool cost; right: reward, over training). All three ungated penalties (flat, OTC, and ungated group-relative) suppress tool use and lose reward; the gated model (green) keeps higher tool use and matches free-tool reward.}
\label{fig:cost_comparison}
\end{figure*}

Finally, we ask whether the model can instead be selective, translating only where the benefit justifies the cost. We compare four cost mechanisms (Figure~\ref{fig:cost_comparison}), applied as reward post-processing on top of the same GSPO setup.

\paragraph{Flat penalty.} Subtracts a fixed cost $\lambda$ from the reward for every tool call regardless of outcome, pressing just as hard on languages that need the tool as on those that do not.

\paragraph{OTC.} The strongest published baseline \citep{wang2025otc} rewards the most tool-efficient response within each group, but still rewards any efficient response, including a no-tool answer that was correct without comprehension.

\paragraph{Ungated group-relative.} The group-relative efficiency method of Nemotron Nano \citep{nemotron2025} applied directly (Section~\ref{sec:continued_rl}): among correct samples in a group, fewer tool calls are rewarded and more are penalized whenever any no-tool answer is correct.

\paragraph{Confidence-Gated GSPO (ours).} Adds a confidence gate to that same signal (Section~\ref{sec:continued_rl}), applying cost pressure only once $K$ of $N$ no-tool samples in the group are correct, so pressure is conditional on demonstrated competence rather than a single correct-by-chance answer. We sweep the gate threshold $K$ and cost strength $\lambda$ and report the best configuration (Appendix~\ref{app:gate_math}).

\begin{table}[t]
\centering
\small
\begin{tblr}{
  colspec = {ll cc},
  hline{1} = {1.5pt, solid},
  hline{Z} = {1.5pt, solid},
  hline{2} = {0.8pt, solid},
  hline{6} = {0.4pt, solid},
  hline{9} = {0.4pt, solid},
  hline{12} = {0.4pt, solid},
  hline{14} = {0.4pt, solid},
  cell{2}{1} = {r=4}{m},
  cell{6}{1} = {r=3}{m},
  cell{9}{1} = {r=3}{m},
  cell{12}{1} = {r=2}{m},
  cell{14}{1} = {r=4}{m},
  row{1} = {font=\bfseries},
  column{1} = {bg=gray!10},
}
  \textbf{Dom} & \textbf{Tier} & Baseline R\% & +RL R\% \\
  Math & High & 63.7 & 66.3 \\
  & Low & 26.2 & 28.0 \\
  & XLow & 40.0 & 42.0 \\
  & Synth & 6.0 & 7.0 \\
  QA & High & 81.0 & 78.8 \\
  & Low & 43.7 & 54.0 \\
  & XLow & 54.0 & 55.0 \\
  IF & High & 72.5 & 78.8 \\
  & Low & 47.3 & 65.0 \\
  & XLow & 55.4 & 71.0 \\
  Summ & High & 43.9 & 52.9 \\
  & Low & 3.4 & 19.0 \\
  Trans & High & 60.0 & 74.0 \\
  & Low & 10.5 & 15.0 \\
  & XLow & 14.7 & 21.0 \\
  & Synth & 3.0 & 7.0 \\
\end{tblr}
\caption{No-tool RL (100 steps). R\% = reward; Baseline = Qwen3-4B zero-shot; +RL = after RL w/o tool access.}
\label{tab:no_tool}
\end{table}

\begin{table*}[t]
\centering
\small
\begin{tblr}{
  width = \textwidth,
  rowsep = 1pt,
  colspec = {ll X[c] X[c] X[c] X[c] X[c] X[c] X[c] X[c] X[c] X[c]},
  hline{1} = {1.5pt, solid},
  hline{Z} = {1.5pt, solid},
  hline{2} = {3-12}{0.4pt, solid},
  hline{3} = {3-12}{0.4pt, solid},
  hline{4} = {0.8pt, solid},
  hline{8} = {0.4pt, solid},
  hline{11} = {0.4pt, solid},
  hline{14} = {0.4pt, solid},
  hline{16} = {0.4pt, solid},
  cell{1}{3} = {c=10}{c},
  cell{2}{3} = {c=2}{c},
  cell{2}{5} = {c=2}{c},
  cell{2}{7} = {c=2}{c},
  cell{2}{9} = {c=2}{c},
  cell{2}{11} = {c=2}{c},
  cell{4}{1} = {r=4}{m},
  cell{8}{1} = {r=3}{m},
  cell{11}{1} = {r=3}{m},
  cell{14}{1} = {r=2}{m},
  cell{16}{1} = {r=4}{m},
  row{1-3} = {m, font=\bfseries},
  column{4,6,8,10,12} = {bg=gray!15},
}
  & & \textbf{Baseline, free, ungated penalties, and gated (ours)} & & & & & & & & & \\
  \textbf{Dom} & \textbf{Tier} & \textbf{Baseline} & & \textbf{Free} & & \textbf{Flat} & & \textbf{OTC} & & \textbf{Gated} & \\
  & & R\% & C\% & R\% & C\% & R\% & C\% & R\% & C\% & R\% & C\% \\
  Math & High & 64.3 & 0.7 & 64.3 & 87.7 & 65.7 & 1.3 & 66.0 & 2.0 & 65.3 & 26.8 \\
  & Low & 30.1 & 7.3 & 62.8 & 92.7 & 51.5 & 27.8 & 42.2 & 22.8 & 61.2 & 55.6 \\
  & XLow & 41.2 & 6.2 & 61.7 & 91.0 & 52.5 & 20.8 & 51.5 & 16.2 & 62.3 & 51.2 \\
  & Synth & 28.0 & 43.0 & 56.0 & 95.0 & 49.0 & 46.0 & 42.0 & 55.0 & 54.0 & 63.5 \\
  QA & High & 82.3 & 0.0 & 82.3 & 77.2 & 82.0 & 0.3 & 82.7 & 0.0 & 80.7 & 19.7 \\
  & Low & 49.3 & 1.4 & 76.0 & 66.2 & 48.7 & 1.4 & 50.2 & 1.5 & 74.5 & 47.5 \\
  & XLow & 50.0 & 1.0 & 66.0 & 63.2 & 54.5 & 0.5 & 58.5 & 1.8 & 64.5 & 47.0 \\
  IF & High & 69.7 & 0.0 & 70.1 & 0.0 & 75.1 & 0.0 & 76.8 & 0.0 & 74.5 & 0.0 \\
  & Low & 48.0 & 0.9 & 60.4 & 2.4 & 61.4 & 0.0 & 59.9 & 0.0 & 61.2 & 0.7 \\
  & XLow & 52.4 & 0.2 & 62.9 & 0.5 & 64.3 & 0.0 & 63.8 & 0.5 & 64.4 & 0.0 \\
  Summ & High & 44.5 & 0.0 & 48.7 & 65.3 & 48.5 & 0.3 & 48.7 & 12.5 & 51.8 & 23.7 \\
  & Low & 2.5 & 31.8 & 13.9 & 94.6 & 9.8 & 13.8 & 10.7 & 82.8 & 12.3 & 89.6 \\
  Trans & High & 78.2 & 11.1 & 89.6 & 52.5 & 90.5 & 50.0 & 91.2 & 50.4 & 89.5 & 50.0 \\
  & Low & 32.1 & 21.2 & 68.1 & 52.2 & 69.5 & 49.8 & 67.7 & 50.5 & 70.2 & 51.1 \\
  & XLow & 31.1 & 19.1 & 52.4 & 55.2 & 54.3 & 50.4 & 56.1 & 50.9 & 53.5 & 51.3 \\
  & Synth & 44.5 & 66.0 & 53.0 & 91.5 & 49.0 & 71.0 & 48.2 & 78.5 & 55.8 & 75.5 \\
\end{tblr}
\caption{Main results: reward and cost by domain and tier. R\% = reward, C\% = tool usage (max 2 calls; gray columns). Baseline = untrained Qwen3-4B; Free = unconstrained tool use (reward upper bound); Flat = fixed per-call penalty; OTC = OTC penalty \citep{wang2025otc}; Gated = our confidence-gated method. All cost mechanisms use $\lambda{=}0.1$. Gated is the only method that trims cost on the high-resource tiers while preserving Free's reward on the low-resource ones; the ungated penalties (Flat, OTC) collapse tool use and lose reward on the tiers that most need it.}
\label{tab:cost_focused}
\end{table*}

\section{Results and Analysis}
\label{sec:results}

\subsection{Performance Gains and Multilingual Ceiling}
\label{sec:no_tool_analysis}
Without any tool, continued RL improves the model on natural languages (Table~\ref{tab:no_tool}), with the clearest gains in instruction following and QA. The consistent improvement across many languages and domains is itself evidence that our translated data is clean. But these gains hit a ceiling on low-resource languages: even after RL, native-only performance stays low in absolute terms (Math Low reaches only 28\%, Translation Low 15\%), well short of the high-resource tiers. The model cannot reason its way out of inputs it does not understand. The synthetic languages are the extreme case of this: with no prior exposure, their scores stay flat through the continued-RL stage, showing the model has no understanding of them at all and cannot solve them without the tool.

\subsection{Unconstrained Tool Use}
\label{sec:free_analysis}
With the tool available and no cost penalty, the model adopts it rapidly and improves reward across tiers, but converges to using it indiscriminately. Since a high-quality translation never lowers reward, nothing pushes back on unnecessary calls: high-resource math climbs to near-total tool use despite no gain over solving natively, and the model settles into calling the tool twice on nearly every input, even where a second call cannot help because the answer is \code{\textbackslash boxed\{42\}} or a single letter identical in any language. A useful signal does appear early in training, with tool use rising first for the low-resource languages that need it most, but without cost pressure this selectivity is fragile and washes out into a blanket ``always translate.''

\subsection{Standard Cost Mechanisms}
\label{sec:cost_analysis}
The main results (Table~\ref{tab:cost_focused}) show that the gate wins exactly where it matters. The ungated penalties, flat and OTC, edge out the gated policy by only 1--2 points on the high-resource tiers (for example, QA High 82.7/82.0 for OTC/flat vs.\ 80.7, IF High 76.8 vs.\ 74.5), precisely the tiers where translation is unnecessary and the ``win'' amounts to refusing a tool that was not needed. But on the low-resource and synthetic tiers, where the tool is essential, the same penalties collapse: on QA Low the gated policy scores 74.5 against 48.7 (flat) and 50.2 (OTC), a 24--26 point gap, and on Math Low 61.2 against 51.5 and 42.2. The ungated methods thus obtain a marginal advantage on tiers that do not require the tool at the cost of 10--26 points on the tiers that do, whereas the gated policy sacrifices little on either. Sweeping the penalty strength $\lambda$ does not rescue them: every setting converges to the same suppressed plateau (Figure~\ref{fig:lambda_strength}). The failure is shared because all the ungated penalties suppress the tool whenever a no-tool answer happens to be correct; the Nemotron Nano group-relative signal collapses the same way when applied without the gate (Appendix~\ref{app:ungated}, Table~\ref{tab:ungated}).

\subsection{Confidence-Gated GSPO}
\label{sec:cascade_analysis}
The gate fixes this by applying cost pressure only once the model has shown it can solve the input on its own: most no-tool attempts in a group must be correct before any penalty applies. For a language it understands, this bar is easily met, so unnecessary translation is trimmed; for one it does not understand, the bar is almost never met, so necessary translation is protected. The gate therefore adapts to each language's difficulty on its own, with no tier labels or hand-set thresholds.

\paragraph{The recovered policy discriminates by language and domain.} Under the gate, high-resource math and QA tool use drops sharply as the model recognizes it can solve them natively, while Low, XLow, and synthetic stay high (per-tier cost in Figure~\ref{fig:tier_comparison}). The domain axis is equally sharp: instruction following stays at zero tool use, translation stays high even for high-resource languages (the task \textit{is} translation, and the 122B translator beats the 4B model's native output), and summarization splits by tier. The policy is also selective \textit{within} a language: its per-prompt tool-use distribution is bimodal (Appendix~\ref{app:bimodal}), calling the tool on problems that need it and skipping those it can solve, rather than applying one rate per language.

\paragraph{Emergent multi-step workflow.} The gate also sharpens discrimination in \textit{how} the tool is used, not just whether. Low-resource summarization requires a multi-step workflow (translate the article in, summarize in English, translate the summary back, format) that is never demonstrated, yet two-call usage and success on the correct translate-in/translate-out pattern both climb over training.

\paragraph{Cost--reward trade-off.} Taken together, this is our central efficiency result: a single policy that recovers full unconstrained reward at a fraction of the cost, without any per-language tuning. The gated policy stays close to the free upper bound on the low-resource, XLow, and synthetic tiers while trimming tool use on the high-resource ones, where the free policy pays for its marginal reward edge with near-total tool use (per-tier cost over training in Appendix~\ref{app:cost_dynamics}).

\subsection{Synthetic Languages}

Kivari and Toqal are the cleanest test of whether the policy tracks genuine comprehension: with no prior exposure the model cannot partially guess its way through them, so the tool is the only route to a correct answer. Two observations confirm the policy reads this correctly. First, native competence is genuinely absent, no-tool RL barely moves synthetic performance, leaving it far below every natural tier, so any real gain must come from the tool. Second, the gated policy responds exactly as it should: it keeps tool usage high even under cost pressure, close to the free model's rate rather than the sharply reduced rate it applies to high-resource languages, and improves +18.7 over the baseline. This is the same behavior it shows on extreme low-resource natural languages, only stronger, so the gate does not treat unseen languages as a special case but places them at the far end of the same competence continuum it has learned to read.

\subsection{Scaling, Unseen Languages, and External Benchmarks}
\label{sec:generalization}

We test whether the learned policy holds beyond its training conditions along three axes: a larger model, unseen languages, and independently built benchmarks. In each setting the gated policy remains the strongest cost-aware method, recovering most of the free-tool reward on the low-resource tiers where the ungated penalties collapse.

\paragraph{Scaling to 8B.} Replicating the free and gated runs on Qwen3-8B under an identical setup (Appendix~\ref{app:8b}, Table~\ref{tab:cost_focused_8b}), the same efficiency holds at larger scale: the gate recovers 96\% of the free-tool gain on Math Low and 94\% on QA Low at 65--76\% of the free tool's cost, and on the high-resource tiers cuts tool use sharply at near-flat reward (QA High from 58.5\% to 9.8\%, Math High from 50.0\% to 22.3\%).

\paragraph{Held-out languages.} On 9 languages absent from training (Table~\ref{tab:generalization}, Appendix~\ref{app:extra_results}), the gate keeps the behavioral signature it learned: it translates less for high- than low-resource languages and never invokes the tool for instruction following. On the low-resource tiers it recovers 85\% of the free-tool gain on Math and 94\% on QA, against at most 60\% (Math) and 12\% (QA) for flat and OTC, which collapse tool use on low resource.

\paragraph{External benchmarks.} On two benchmarks we did not build, MGSM \citep{shi2023language} and Global MMLU \citep{globalmmlu2024} (Table~\ref{tab:external}, Appendix~\ref{app:extra_results}), the pattern holds off our own data: the gate recovers 91\% of the free-tool gain on MGSM Low and 80\% on MMLU Low at 60--78\% of the free tool's cost, against at most 40\% (MGSM) and 6\% (MMLU) for flat and OTC, which leave reward stranded near the untrained baseline.

\section{Conclusion}

We presented a single policy that learns \textit{when} to translate from task reward alone. Continuing RL on Qwen3-4B across 22 languages and 5 domains, it develops language- and domain-adaptive introspection, translating only when it cannot solve a task natively, with no language labels or routing rules. Our confidence-gated GSPO makes this behavior cost-sensitive, preserving full unconstrained reward at 66\% of the cost: it outperforms standard flat and OTC cost penalties on the low-resource tiers while matching them on the high-resource tiers where the tool is unnecessary. Underpinning this, our answer-preserving translation pipeline supplies clean multilingual reward at 98.4\% fidelity, the core requirement for RLVR on translated data. Finally, on synthetic languages with zero prior exposure, the policy correctly learns to always translate, showing it assesses genuine comprehension rather than surface cues. The learned behavior is not tied to the training setup: it transfers zero-shot to 9 held-out languages and holds on the native MGSM and Global MMLU benchmarks.
\section*{Limitations}

\paragraph{Translation model quality.} Our pipeline relies on Qwen3.5-122B-A10B as translator, which exhibits language proximity confusion for some low-resource languages (defaulting to a related higher-resource neighbor). Our final evaluation covers 22 natural languages where translation accuracy exceeds 98\%.

\paragraph{Policy--translator size ratio.} A 4B policy calling a 122B translator is not a realistic deployment ratio. We use the larger model deliberately: it holds translation quality high, particularly for the low-resource languages where weaker translators are error-prone, so that what we measure is the policy's when-to-translate decision rather than translator noise. In a real deployment the 122B translator would be replaced by a smaller, dedicated, more competent translation model or service behind the same tool interface, leaving the learned policy unchanged.

\paragraph{Human evaluation.} We do not conduct human evaluation. For quality checks that would ideally use human annotators, such as auditing translation fidelity across low-resource languages, we use Claude Opus as a proxy annotator. While this scales to all our languages and agrees closely with our own manual spot-checks, it inherits the biases and blind spots of an automated judge and is not a substitute for native-speaker evaluation.

\paragraph{LLM-judge variance.} Rewards for the non-verifiable domains (summarization, translation) come from an LLM judge rather than a deterministic check. This signal carries more variance than the exact-match rewards used for math and QA, which contributes to the more modest and noisier results we observe on those domains.

\paragraph{Resource tiers are approximate.} We assign tiers by approximate digital-speaker count, a proxy for pretraining representation. This does not perfectly predict per-task competence: some XLow languages outperform Low ones on specific tasks, likely through script, vocabulary, or proximity overlap with high-resource languages. Our method does not depend on these labels, since the confidence gate adapts to difficulty without tier information, but the tier groupings in our tables should be read as coarse rather than exact.

\section*{Broader Impact}

Selective translation can broaden access to LLM capabilities for speakers of low-resource languages by routing comprehension through a stronger language only when needed, reducing unnecessary compute for languages the model already handles. However, reliance on a translation model inherits its biases: translation errors or cultural flattening may propagate into downstream answers, and quality remains uneven across languages. The translation pipeline also carries a compute footprint (a large translator served alongside the policy), which deployments should weigh against the accessibility benefits.

\bibliography{custom}

\appendix

\begin{center}
\Large\textbf{Appendix}
\end{center}
\vspace{8pt}

\section{Data Sources}
\label{app:data_sources}

We draw each domain from an established English dataset and translate it into our 22 target languages through the pipeline of Section~\ref{sec:pipeline}; summarization instead uses native-language text where it exists. Table~\ref{tab:data_sources} lists the source for each domain.

\begin{table*}[t]
\centering
\small
\begin{tblr}{
  width = \textwidth,
  colspec = {l l X[l]},
  hline{1} = {1.5pt, solid},
  hline{Z} = {1.5pt, solid},
  hline{2} = {0.8pt, solid},
  row{1} = {font=\bfseries},
  column{1} = {bg=gray!10},
}
  Domain & English Source & Notes \\
  Math & Massive-Math-455K-Verified & HuggingFace; word problems with numeric answers \\
  QA & Nemotron-CrossThink & 187K MCQ with rule-based verification \\
  IF & Custom constraint generation & Keywords + structural constraints; see Appendix~\ref{app:prompts} \\
  Summarization & XL-Sum \citep{xlsum2021} & Native articles in 13 languages (no translation) \\
  Translation & XL-Sum English highlights & Bidirectional pairs via 5-stage pipeline \\
\end{tblr}
\caption{English source datasets per domain. All verifiable domains (Math, QA, IF) are translated to 22 target languages via our pipeline; Summarization uses native data where available.}
\label{tab:data_sources}
\end{table*}

\section{Data Distribution}
\label{app:data_dist}

We weight the training mix inversely to resource level, giving low-resource languages several times more samples per language than high-resource ones so the reward signal is not dominated by languages the model already handles. Tables~\ref{tab:data_dist_v} and~\ref{tab:data_dist} break down the 273K training samples by tier for the verifiable and non-verifiable domains.

\begin{table}[t]
\centering
\small
\begin{tblr}{
  colspec = {l cccc},
  hline{1} = {1.5pt, solid},
  hline{Z} = {1.5pt, solid},
  hline{2} = {0.8pt, solid},
  hline{7} = {0.4pt, solid},
  row{1} = {font=\bfseries},
  row{Z} = {font=\bfseries},
  column{1} = {bg=gray!10},
}
  Tier & Math & QA & IF & Total \\
  High (6) & 2.9K & 2.9K & 2.9K & 8.7K \\
  Low (12) & 63.0K & 58.2K & 63.8K & 185.0K \\
  XLow (4) & 16.8K & 11.3K & 17.8K & 45.9K \\
  Synth (2) & 4.0K & -- & -- & 4.0K \\
  Total & 87K & 72K & 85K & 244K \\
\end{tblr}
\caption{Verifiable training data by tier.}
\label{tab:data_dist_v}
\end{table}

\begin{table}[t]
\centering
\small
\begin{tblr}{
  colspec = {l ccc},
  hline{1} = {1.5pt, solid},
  hline{Z} = {1.5pt, solid},
  hline{2} = {0.8pt, solid},
  hline{7} = {0.4pt, solid},
  row{1} = {font=\bfseries},
  row{Z} = {font=\bfseries},
  column{1} = {bg=gray!10},
}
  Tier & Summ & Trans & Total \\
  High (6) & 1.9K & 0.6K & 2.5K \\
  Low (12) & 11.3K & 11.6K & 22.9K \\
  XLow (4) & -- & 1.7K & 1.7K \\
  Synth (2) & -- & 2.0K & 2.0K \\
  Total & 13K & 16K & 29K \\
\end{tblr}
\caption{Non-verifiable training data by tier. Combined with Table~\ref{tab:data_dist_v}: 273K total.}
\label{tab:data_dist}
\end{table}

\begin{table}[t]
\centering
\small
\begin{tblr}{
  colspec = {l c X[l]},
  hline{1} = {1.5pt, solid},
  hline{Z} = {1.5pt, solid},
  hline{2} = {0.8pt, solid},
  row{1} = {font=\bfseries},
  column{1} = {bg=gray!10},
}
  Tier & N & Languages \\
  High & 6 & English, Arabic, Chinese, French, Japanese, Russian \\
  Low & 12 & Amharic, Aymara, Guarani, Hausa, Kinyarwanda, Luganda, Quechua, Somali, Uyghur, Wolof, Yoruba, Zulu \\
  XLow & 4 & Bambara, Ewe, Lingala, Twi \\
\end{tblr}
\caption{The 22 natural languages by resource tier, assigned by approximate digital-speaker count (Section~\ref{sec:pipeline}).}
\label{tab:languages}
\end{table}

\section{Translation Pipeline Details}
\label{app:pipeline_details}

\paragraph{Source Filtering Thresholds.} For math, we retain only word problems where $\geq$80\% of content is natural language and total length is under 2,000 characters. This eliminates approximately 40\% of raw source data. For QA, we filter by option count (must have A/B/C/D) and question length bounds.

\paragraph{Forward Translation Prompts.} Each domain has a dedicated translation prompt. For math: ``Keep all mathematical notation (LaTeX, equations, numbers, variable names) exactly as-is. Only translate the natural language parts.'' For QA: structured JSON input preserving option keys. Output wrapped in \code{<output>} tags.

\paragraph{Heuristic Thresholds.}
\begin{itemize}
    \item Character n-gram repetition: reject if repetition ratio $>$ 0.3 (n=10)
    \item Length ratio: reject if translation/source ratio $<$ 0.2 or $>$ 3.0
    \item Source copy: reject if chrF(translation, source) $>$ 0.9
    \item Translation domain: max tokens 2048 (math/QA), 512 (translation)
\end{itemize}

\paragraph{Back-Translation chrF Threshold.} chrF $\geq$ 0.4 for math/QA, $\geq$ 0.35 for translation (lower because short sentences have higher chrF variance). Conservative threshold catches substantial content loss before the expensive judge call.

\paragraph{LLM Judge Criteria.} The judge compares original English with back-translation. It explicitly ignores: name/story context changes, rephrasing, format differences, word order. It flags: number changes, operation changes, constraint changes, information loss, answer leakage, definition changes, question changes. Binary verdict: SAME/DIFFERENT. Three retries on bad format; samples failing all retries are rejected.

\paragraph{Concurrency and Quotas.} Translation runs asynchronously with 128--256 concurrent requests per language. Per-language quotas ensure inverse resource weighting (High: 610, Low: 4880, XLow: 3660 samples per domain). Failed samples recycle with replacement from the source pool; 10 attempts maximum per slot before fallback.

\section{Non-Verifiable Domain Data}
\label{app:nonverifiable}

For summarization, we use native-language articles from XL-Sum where available (13 languages), and translate only for languages lacking native data using a fail-fast pipeline: translate the short label first, then the expensive article only if the label passes heuristic and chrF checks, followed by a cross-coherence judge verifying that the back-translated label still summarizes the back-translated article.

Translation data is sourced from XL-Sum English highlights, translated to target languages with the standard five-stage pipeline (heuristic, back-translation, chrF $\geq$ 0.35, LLM meaning-preservation judge), then formatted as bidirectional pairs (English-to-target and target-to-English).

\section{Translation Model Choice}
\label{app:translator}

We use Qwen3.5-122B-A10B-FP8 as translator because no single compact model covers all 22 target languages with sufficient quality. Existing open-source translation models (NLLB \citep{nllb2022}, Gemma-based translators) perform well on high-resource pairs but degrade severely on our lowest-resource targets. In our evaluation, Gemma-based translation models produced near-unusable output for XLow languages like Bambara, Ewe, and Twi, where even basic meaning preservation failed.

While deploying a 122B model alongside a 4B policy model may appear computationally disproportionate, the 122B serves as a proxy for the specialized, efficient translation service that would be used in deployment. Integrating an external API introduced latency and cost constraints incompatible with our synchronous RL training loop, making self-hosted inference the practical choice. One might ask why not use the 122B directly to solve the problems; the answer is that our goal is to develop a policy that knows when it needs help.

\section{Synthetic Language Construction}
\label{app:synthetic}

We build Kivari and Toqal by deterministically substituting every English word with a unique sequence of consonant-vowel-consonant nonsense syllables, using a 962,531-word vocabulary drawn from all our English sources and two independent seeds. The mapping is bijective and length-matched, and numbers, LaTeX commands, and mathematical symbols pass through unchanged so answer extraction still works. We substitute only the prompt, leaving labels intact, and give the languages plausible-sounding names so the model cannot shortcut from the name alone. Each language contributes 2,000 training and 50 evaluation samples for math and translation, drawn from the same English pool as the natural languages.

\section{Data Quality Audit Methodology}
\label{app:audit}

We audit the full evaluation set (up to 50 per language $\times$ 3 verifiable domains $\times$ 16 Low+XLow languages = 2,289 samples) using Claude Opus as annotator. Samples are split into 4 batches and annotated in parallel. Each sample is assessed for: (1) language correctness (is the prompt in the claimed language?), (2) label format (does it match the domain?), (3) content coherence (does original\_en relate to the translation?). The following prompt is used:

\vspace{4pt}
\begin{promptbox}[Data Quality Audit Prompt (Claude Opus)]
\rmfamily\footnotesize
Read the file. It contains one batch (\textasciitilde570) of the 2,289 multilingual eval samples. Each has: domain (math/qa/translation), language, prompt, label, original\_en.\par\medskip
These are translated from English (original\_en) to the target language (prompt). The label is the correct answer.\par\medskip
IMPORTANT:\par
- ``kivari'' and ``toqal'' are INTENTIONAL synthetic cipher languages. They look like random syllables. This is BY DESIGN. Do NOT flag.\par
- Prompts truncated at 500 chars, labels at 300. Don't flag truncation.\par
- Guarani/Lingala may use Spanish/French due to regional bilingualism. Note but not necessarily errors.\par\medskip
Read EVERY sample. For each, check:\par
1. Is the prompt in the claimed language? (Check script: Amharic=Ethiopic, Uyghur=Arabic, etc.)\par
2. Does label format match domain? (math=number/expression, qa=single letter A/B/C/D, translation=text)\par
3. Does original\_en relate to the prompt content?\par
4. For QA: does the label letter seem plausible given the question?\par\medskip
Report ONLY: Total reviewed, number with definite issues, for each issue: language, domain, line number, and what's wrong (1 sentence). Overall pass rate percentage.\par\medskip
Be strict on wrong-language but fair on bilingual regions.
\end{promptbox}

\vspace{4pt}
The audit confirms \textbf{98.4\% fidelity} across all Low and XLow languages.

\section{Evaluation Data}
\label{app:eval_data}

Each language has up to 50 evaluation samples per domain. Table~\ref{tab:eval_samples} shows the total evaluation samples per tier.

\begin{table*}[t]
\centering
\small
\begin{tblr}{
  colspec = {l *{6}{r}},
  hline{1} = {1.5pt, solid},
  hline{Z} = {1.5pt, solid},
  hline{2} = {0.8pt, solid},
  row{1} = {font=\bfseries},
  row{Z} = {font=\bfseries},
  column{1} = {bg=gray!10},
}
  Tier & Math & QA & IF & Summ & Trans & Total \\
  High (6) & 300 & 300 & 300 & 300 & 140 & 1,340 \\
  Low (12) & 600 & 600 & 600 & 250 & 574 & 2,624 \\
  XLow (4) & 200 & 200 & 200 & -- & 115 & 715 \\
  Synth (2) & 100 & -- & -- & -- & 100 & 200 \\
  Total & 1,200 & 1,100 & 1,100 & 550 & 929 & 4,879 \\
\end{tblr}
\caption{Evaluation samples per tier and domain.}
\label{tab:eval_samples}
\end{table*}

\section{Hyperparameters}
\label{app:hyperparams}

Table~\ref{tab:hyperparams} lists the training hyperparameters, which are held fixed across all RL variants (no-tool, free, gated, flat, OTC) so that differences between them come only from the cost mechanism. We use GSPO with 8 samples per prompt, a constant learning rate, and no KL penalty; the tool runs train for 250 rollout steps and the no-tool run for 100.

\begin{table}[t]
\centering
\small
\begin{tblr}{
  colspec = {l l},
  hline{1} = {1.5pt, solid},
  hline{Z} = {1.5pt, solid},
  hline{2} = {0.8pt, solid},
  row{1} = {font=\bfseries},
  column{1} = {bg=gray!10},
}
  Parameter & Value \\
  Learning rate & 1e-6 (constant) \\
  Prompts per rollout & 768 \\
  Samples per prompt & 8 \\
  Total rollout steps & 250 (tool runs), 100 (no-tool) \\
  Eval frequency & Every 10 steps \\
  Max generation length & 20{,}000 tokens \\
  Temperature & 1.0 \\
  Top-p & 1.0 \\
  KL penalty & None \\
\end{tblr}
\caption{Training hyperparameters.}
\label{tab:hyperparams}
\end{table}

\section{Gate Threshold and Cost Strength}
\label{app:gate_math}

We sweep the gate threshold $K \in \{4, 6\}$ and the cost strength $\lambda$, early-stopping runs once their behavior is clear, and report the best configuration.

\paragraph{Threshold $K$.} The threshold sets how much native competence the gate demands before applying cost pressure. A lenient $K{=}4$ still suppresses tool use on exactly the low-resource tiers that need it: at a step-matched checkpoint, its average tool cost on the low-resource and synthetic Math and QA tiers is only 29.9\%, versus 41.6\% for $K{=}6$, and the suppression is sharpest where the tool matters most (for example, Math Low tool use is 38\% for $K{=}4$ against 58\% for $K{=}6$, and Math XLow 32\% against 54\%). $K{=}4$ therefore leaves the cascade partially intact, so we use $K{=}6$.

\paragraph{Cost strength $\lambda$.} As with the ungated mechanisms (Figure~\ref{fig:lambda_strength}), the cost strength has little effect on the outcome: sweeping $\lambda$ does not change the qualitative behavior, so we simply keep $\lambda{=}0.1$.

\section{Cost Mechanism Comparison}
\label{app:ungated}

The main results table (Table~\ref{tab:cost_focused}) already compares the flat and OTC penalties against our gated policy. Here we isolate the effect of the confidence gate itself: Table~\ref{tab:ungated} places the ungated group-relative efficiency signal of Nemotron Nano \citep{nemotron2025}, the mechanism our gate builds on, next to our gated policy that adds the gate to it, with the unconstrained free policy as a reward upper bound. All use $\lambda{=}0.1$ and the same setup as Table~\ref{tab:cost_focused}. Removing the gate collapses the mechanism onto the same failure mode as the other ungated penalties: on the low-resource Math and QA tiers the ungated signal suppresses tool use (Math Low cost 26.8\%, QA Low 2.2\%) and loses the reward the tool was there to provide (QA Low 51.3 versus 74.5 for the gated policy and 76.0 for free), whereas the gated policy retains it. This confirms that the gate, not the group-relative efficiency signal itself, is what preserves low-resource reward.

\begin{table*}[t]
\centering
\small
\begin{tblr}{
  width = \textwidth,
  rowsep = 1pt,
  colspec = {ll X[c] X[c] X[c] X[c] X[c] X[c]},
  hline{1} = {1.5pt, solid},
  hline{Z} = {1.5pt, solid},
  hline{2} = {3-8}{0.4pt, solid},
  hline{3} = {3-8}{0.4pt, solid},
  hline{4} = {0.8pt, solid},
  hline{8} = {0.4pt, solid},
  hline{11} = {0.4pt, solid},
  hline{14} = {0.4pt, solid},
  hline{16} = {0.4pt, solid},
  cell{1}{3} = {c=6}{c},
  cell{2}{3} = {c=2}{c},
  cell{2}{5} = {c=2}{c},
  cell{2}{7} = {c=2}{c},
  cell{4}{1} = {r=4}{m},
  cell{8}{1} = {r=3}{m},
  cell{11}{1} = {r=3}{m},
  cell{14}{1} = {r=2}{m},
  cell{16}{1} = {r=4}{m},
  row{1-3} = {m, font=\bfseries},
  column{4,6,8} = {bg=gray!15},
}
  & & \textbf{No-gate ablation} & & & & & \\
  \textbf{Dom} & \textbf{Tier} & \textbf{Free} & & \textbf{Ungated} & & \textbf{Gated (ours)} & \\
  & & R\% & C\% & R\% & C\% & R\% & C\% \\
  Math & High & 64.3 & 87.7 & 67.7 & 0.5 & 65.3 & 26.8 \\
  & Low & 62.8 & 92.7 & 49.2 & 26.8 & 61.2 & 55.6 \\
  & XLow & 61.7 & 91.0 & 53.0 & 16.5 & 62.3 & 51.2 \\
  & Synth & 56.0 & 95.0 & 48.0 & 46.5 & 54.0 & 63.5 \\
  QA & High & 82.3 & 77.2 & 81.3 & 0.0 & 80.7 & 19.7 \\
  & Low & 76.0 & 66.2 & 51.3 & 2.2 & 74.5 & 47.5 \\
  & XLow & 66.0 & 63.2 & 57.5 & 1.0 & 64.5 & 47.0 \\
  IF & High & 70.1 & 0.0 & 74.9 & 0.0 & 74.5 & 0.0 \\
  & Low & 60.4 & 2.4 & 61.3 & 0.0 & 61.2 & 0.7 \\
  & XLow & 62.9 & 0.5 & 64.2 & 0.0 & 64.4 & 0.0 \\
  Summ & High & 48.7 & 65.3 & 50.9 & 2.7 & 51.8 & 23.7 \\
  & Low & 13.9 & 94.6 & 9.5 & 61.0 & 12.3 & 89.6 \\
  Trans & High & 89.6 & 52.5 & 91.1 & 50.0 & 89.5 & 50.0 \\
  & Low & 68.1 & 52.2 & 68.9 & 50.8 & 70.2 & 51.1 \\
  & XLow & 52.4 & 55.2 & 59.1 & 50.9 & 53.5 & 51.3 \\
  & Synth & 53.0 & 91.5 & 37.0 & 56.5 & 55.8 & 75.5 \\
\end{tblr}
\caption{No-gate ablation ($\lambda{=}0.1$). R\% = reward, C\% = tool usage (max 2 calls; gray columns). Free = unconstrained tool use (reward upper bound); Ungated = the group-relative efficiency signal without the confidence gate; Gated = our confidence-gated method. Removing the gate collapses tool use and reward on the low-resource Math and QA tiers, whereas the gated policy retains reward and tracks the free upper bound. Flat and OTC penalties are in Table~\ref{tab:cost_focused}.}
\label{tab:ungated}
\end{table*}

\section{Full System Prompts}
\label{app:prompts}

Each domain uses a task-specific system prompt prepended to the user's input. All prompts are language-agnostic: they instruct the model to handle input ``in any language'' without specifying which one, letting the model decide whether to solve natively or invoke the translation tool. When \code{TOOL\_MODE=free}, the tool-use prompt (bottom-right) is appended, giving the model access to a translate function it may call up to twice per turn. Placeholder variables (e.g., \code{\{language\}}, \code{\{source\_lang\}}) are filled at data-loading time from the sample metadata.

\vspace{4pt}

\begin{promptbox}[Math]
You are a math assistant. You will receive a math problem that may be written in any language.\par\medskip
Instructions:\par
1. Solve the problem step by step, showing your reasoning.\par
2. Place your final answer inside \textbackslash boxed\{\}. The answer inside \textbackslash boxed\{\} should be the mathematical answer only.\par\medskip
Examples of correct output format:\par
- For a numeric answer: \textbackslash boxed\{42\}\par
- For a fraction: \textbackslash boxed\{\textbackslash frac\{1\}\{2\}\}\par
- For an expression: \textbackslash boxed\{2x + 3\}
\end{promptbox}
\vspace{1pt}
\begin{promptbox}[QA (Multiple Choice)]
You are a knowledgeable assistant. You will receive a multiple-choice question that may be written in any language.\par\medskip
Instructions:\par
1. Think through the problem step by step.\par
2. Place your final answer inside <answer> tags. The answer should be ONLY the option letter (A, B, C, D, etc.).\par\medskip
Example output format:\par
I think the answer is B because...\par
<answer>B</answer>
\end{promptbox}
\vspace{1pt}
\begin{promptbox}[Instruction Following]
You are a writing assistant. You will receive a prompt in \{language\} containing:\par
- Keywords: A comma-separated list of words you MUST include.\par
- Constraints: One or more rules your response MUST satisfy.\par\medskip
Instructions:\par
1. Read the keywords and constraints carefully.\par
2. Write a coherent response that naturally incorporates ALL keywords.\par
3. Your response MUST satisfy ALL constraints exactly.\par
4. Write your entire response in \{language\}.\par
5. Place your response inside <answer> tags.
\end{promptbox}
\vspace{1pt}
\begin{promptbox}[Summarization]
You are a summarization assistant. You will receive a news article that may be written in any language.\par\medskip
Instructions:\par
1. Read the article carefully.\par
2. Write a concise summary in EXACTLY 1 sentence that captures the main topic and key details.\par
3. The summary MUST be written in the same language as the article.\par
4. Place your final summary inside <answer> tags.
\end{promptbox}
\vspace{1pt}
\begin{promptbox}[Translation]
You are a translation assistant. Translate the given text from \{source\_lang\} to \{target\_lang\}.\par\medskip
Instructions:\par
1. Translate the text accurately and fluently.\par
2. Preserve the meaning, tone, and style of the original.\par
3. Do not add, remove, or change any information.\par
4. Place your final translation inside <answer> tags.
\end{promptbox}
\vspace{1pt}
\begin{promptbox}[Tool Use (appended when available)]
You have access to a translation tool. To use it, output:\par
<tool\_call>\par
\{"name": "translate", "arguments": \{"text": "...", "target\_lang": "..."\}\}\par
</tool\_call>\par\medskip
The tool will respond with the translation:\par
<tool\_response>\par
<translated text>\par
</tool\_response>\par\medskip
You can translate to or from any language.\par
You may use the tool up to 2 times, or not at all.\par
For example, you can translate the input to English to understand it, then translate your answer back.
\end{promptbox}

\vspace{6pt}
\captionof{figure}{System prompts for all five domains plus the tool-use prompt appended when \code{TOOL\_MODE=free}.}
\label{fig:system_prompts}

\section{Full Reward Judge Prompts}
\label{app:judge_prompts}

Math and QA use deterministic extraction (\code{\textbackslash boxed\{\}} matching and option-letter parsing) and do not require an LLM judge. The remaining three domains use an LLM judge (Qwen3.5-122B-A10B) that evaluates the model's response against the reference answer, with the judge prompts shown below. Summarization applies heuristic pre-filters (verbatim copying, length ratio) before the LLM call to avoid wasting judge compute on clearly incorrect outputs. All judges output structured scores inside XML tags for reliable parsing.

\vspace{4pt}

\begin{promptbox}[Summarization Judge]
\rmfamily\textit{Pre-filters (before LLM call):}\par
1. Verbatim gate: If $>$50\% of sentences copied from source $\to$ 0\par
2. Length penalty: $\text{mult} = e^{-0.5(r-2)}$ for ratio $r = \text{generated length} / \text{reference length} > 2$\par\medskip
\rmfamily\textit{LLM judge prompt:}\par\medskip
\ttfamily
You are an expert summarization evaluator.\par\medskip
Gates (PASS/FAIL; if ANY fails, score is 0):\par
1. Main Topic: Same topic as reference?\par
2. Factual Accuracy: No hallucinations?\par
3. Is a Summary: Condensed, not a copy?\par\medskip
Score (only if all gates pass):\par
4. Key Detail Coverage: (0/1/2/3/4)\par\medskip
Output: <output>FAIL</output>\par
or: <output>PASS,3</output>\par\medskip
\rmfamily\textit{Final reward:} $\text{score}/4 \times \text{length\_mult}$
\end{promptbox}
\vspace{1pt}
\begin{promptbox}[Translation Judge]
You are an expert translation evaluator.\par\medskip
Score on two dimensions (0, 1, or 2 each):\par
1. Accuracy: Does the translation preserve the meaning of the original?\par
2. Completeness: Is all content translated without omission?\par\medskip
Output ONLY inside <output> tags: two scores separated by commas.\par
Example: <output>2,1</output>\par\medskip
\rmfamily\textit{Final reward:} $(\text{accuracy} + \text{completeness})/4$
\end{promptbox}
\vspace{1pt}
\begin{promptbox}[Instruction Following Judge]
Evaluate on two dimensions (0, 1, or 2 each):\par
1. Language: Written in \{language\}? (0=English, 1=mixed, 2=target)\par
2. Coherence: Fluent? (0=gibberish, 1=partial, 2=fluent)\par\medskip
Output: <output>2,2</output>\par\medskip
\rmfamily\textit{Final reward:} $\frac{\text{lang}+\text{coh}}{4} \times (0.5 \cdot \text{kw} + 0.5 \cdot \text{cstr})$\par
Language = 0 is a hard gate (reward = 0).
\end{promptbox}

\vspace{6pt}
\captionof{figure}{Reward judge prompts for summarization, translation, and instruction following.}
\label{fig:judge_prompts}

\section{Per-Problem Tool-Use Distribution}
\label{app:bimodal}

\begin{figure}[htbp]
\centering
\includegraphics[width=\columnwidth]{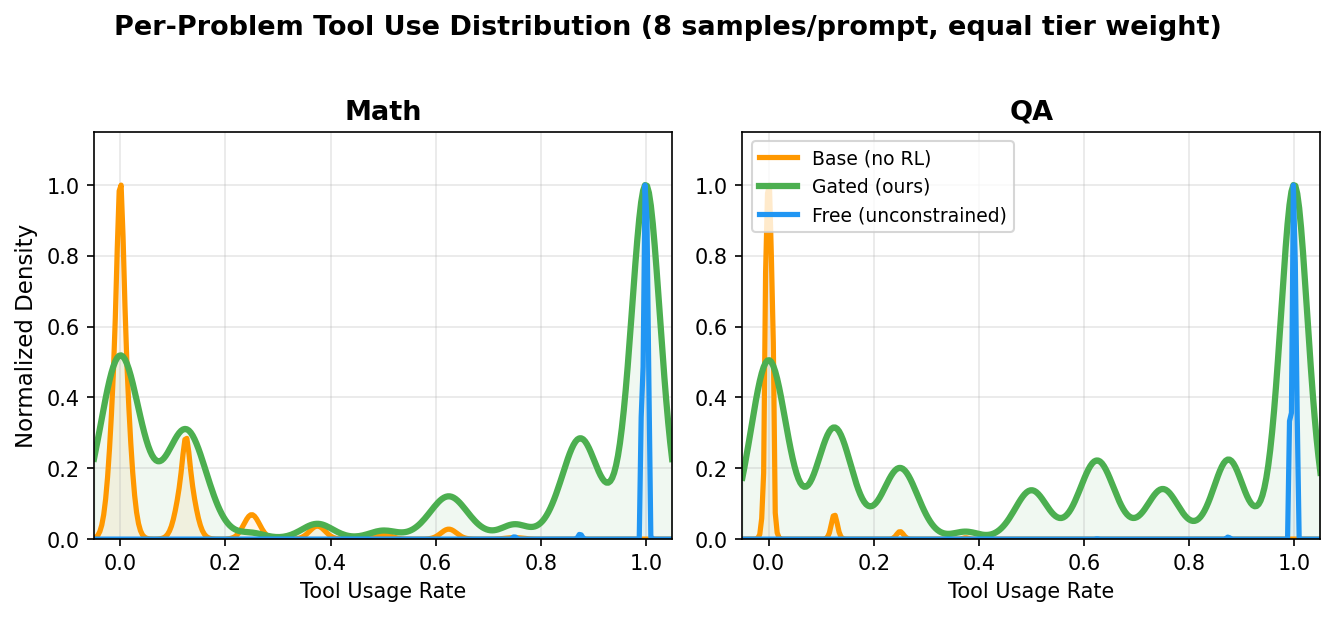}
\caption{Per-problem tool use distribution (8 samples/prompt, equal tier weighting). Baseline (orange) clusters at 0, Free (blue) at 1, Gated (green) shows bimodal selectivity with peaks at both extremes.}
\label{fig:bimodal}
\end{figure}

Figure~\ref{fig:bimodal} plots the distribution of per-prompt tool-use rates (8 samples each, equal tier weighting), supporting the per-problem selectivity discussion in Section~\ref{sec:cascade_analysis}.

\section{Cost Mechanism Dynamics}
\label{app:cost_dynamics}

\begin{figure}[htbp]
\centering
\includegraphics[width=\columnwidth]{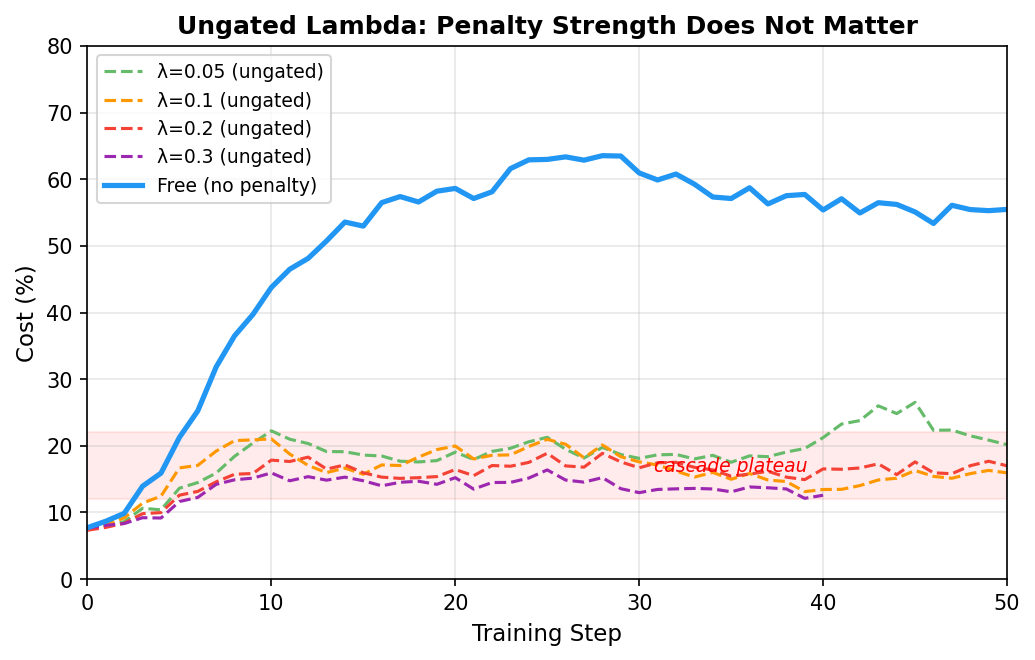}
\caption{Penalty strength does not matter: all ungated $\lambda$ values (0.05--0.3) converge to the same cascade plateau.}
\label{fig:lambda_strength}
\end{figure}

\begin{figure*}[htbp]
\centering
\includegraphics[width=\textwidth]{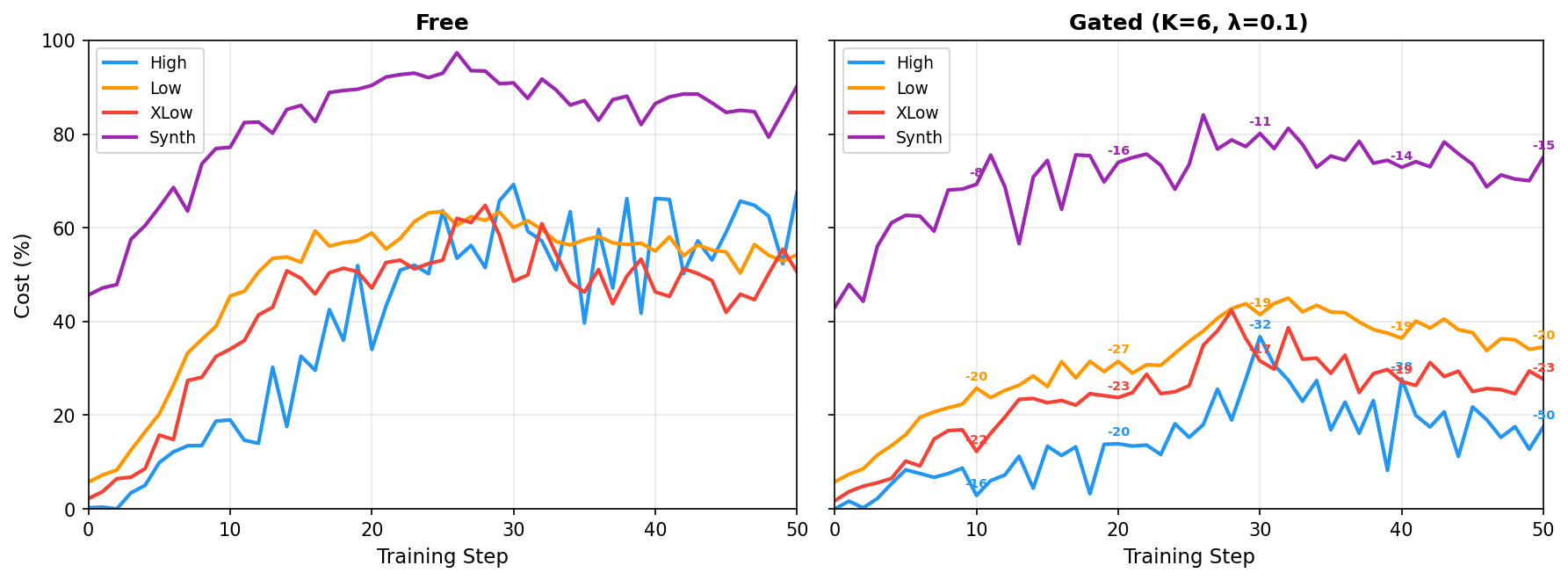}
\caption{Per-tier cost over training. Left: free-tool converges to uniformly high cost. Right: gated keeps tier separation, trimming High-resource cost while preserving Low/Synth adoption.}
\label{fig:tier_comparison}
\end{figure*}

These figures support the cost-mechanism analysis of Section~\ref{sec:cost_sensitive}. Figure~\ref{fig:lambda_strength} shows that varying the penalty strength $\lambda$ does not change the ungated outcome, and Figure~\ref{fig:tier_comparison} shows how per-tier cost evolves over training for the free and gated policies.

\section{Transfer Results}
\label{app:extra_results}

This section gives the full per-tier numbers behind the transfer analysis of Section~\ref{sec:generalization}: zero-shot transfer to held-out languages (Table~\ref{tab:generalization}) and results on the external MGSM and Global MMLU benchmarks (Table~\ref{tab:external}).

\paragraph{Held-out languages.} A central claim of this work is that the model learns a general \emph{when-to-translate} policy rather than memorizing behavior for the 22 languages it trained on. To test this directly, we evaluate all five policies zero-shot on 9 languages that appear nowhere in training (High: Hindi, Turkish, Korean; Low: Igbo, Shona, Sesotho, Oromo, Tswana, Maori), across the three verifiable domains where reward is directly comparable across languages (Table~\ref{tab:generalization}). If the policy had merely fit per-language rules, it would have no basis for acting sensibly here; instead every qualitative behavior from the in-training results reappears. On Math, the tier gap the model learned to exploit transfers cleanly: it drives Low-resource reward from a 23.7 baseline to 58.0 while calling the tool on only 56\% of Low inputs but 32\% of High inputs, mirroring the language-adaptive discrimination seen on trained languages. On QA it is even sharper, lifting Low from 43.0 to 73.7 and, notably, achieving the best High-resource reward of any policy (82.7) at just 26\% tool use, meaning it recovers essentially all of the free model's benefit while translating far less. Instruction following behaves exactly as in training: tool use stays at or near zero across both tiers because the bottleneck is generation rather than comprehension, and the gated policy still posts the top reward (72.3 High, 57.7 Low). The contrast with the ungated baselines is the same one we see on trained languages: flat and OTC collapse Low-resource Math tool use to 32\% and 26\% and lose most of the reward (48.0 and 37.3, versus 58.0 for gated), confirming that their failure mode is structural and not an artifact of the training languages. Taken together, the policy transfers as a principle, translate when native competence is weak and abstain when it is not, rather than as a lookup over seen languages.

\begin{table*}[t]
\centering
\small
\begin{tblr}{
  width = \textwidth,
  colspec = {ll X[c] X[c] X[c] X[c] X[c] X[c] X[c] X[c] X[c] X[c]},
  hline{1} = {1.5pt, solid},
  hline{Z} = {1.5pt, solid},
  hline{2} = {3-12}{0.4pt, solid},
  hline{3} = {3-12}{0.4pt, solid},
  hline{4} = {0.8pt, solid},
  hline{6} = {0.4pt, solid},
  hline{8} = {0.4pt, solid},
  cell{1}{3} = {c=10}{c},
  cell{2}{3} = {c=2}{c},
  cell{2}{5} = {c=2}{c},
  cell{2}{7} = {c=2}{c},
  cell{2}{9} = {c=2}{c},
  cell{2}{11} = {c=2}{c},
  cell{4}{1} = {r=2}{m},
  cell{6}{1} = {r=2}{m},
  cell{8}{1} = {r=2}{m},
  row{1-3} = {m, font=\bfseries},
  column{4,6,8,10,12} = {bg=gray!15},
}
  & & \textbf{Held-Out Languages (Zero-Shot Transfer)} & & & & & & & & & \\
  \textbf{Dom} & \textbf{Tier} & \textbf{Baseline} & & \textbf{+RL Free} & & \textbf{+RL Flat} & & \textbf{+RL OTC} & & \textbf{+RL Gate} & \\
  & & R\% & C\% & R\% & C\% & R\% & C\% & R\% & C\% & R\% & C\% \\
  Math & High & 56.7 & 0 & 62.0 & 87 & 58.7 & 2 & 56.7 & 2 & 59.3 & 32 \\
  & Low & 23.7 & 8 & 64.0 & 94 & 48.0 & 32 & 37.3 & 26 & 58.0 & 56 \\
  QA & High & 80.0 & 0 & 80.0 & 72 & 78.7 & 0 & 82.0 & 0 & 82.7 & 26 \\
  & Low & 43.0 & 2 & 75.7 & 66 & 44.7 & 1 & 47.0 & 1 & 73.7 & 50 \\
  IF & High & 61.4 & 0 & 68.1 & 1 & 70.7 & 0 & 72.5 & 0 & 72.3 & 0 \\
  & Low & 50.9 & 1 & 57.4 & 2 & 57.4 & 0 & 57.8 & 0 & 57.7 & 0 \\
\end{tblr}
\caption{Zero-shot transfer to 9 held-out languages (High: Hindi, Turkish, Korean; Low: Igbo, Shona, Sesotho, Oromo, Tswana, Maori), none seen in training. R\% = reward, C\% = tool cost.}
\label{tab:generalization}
\end{table*}

\paragraph{External benchmarks.} A reasonable worry is that our results are specific to data produced by our own translation pipeline: the model might be exploiting artifacts of how we translate rather than a genuine comprehension signal. To rule this out, we evaluate on two established, natively multilingual benchmarks that we did not construct, MGSM \citep{shi2023language} for math and Global MMLU \citep{globalmmlu2024} for knowledge QA (Table~\ref{tab:external}). We split each into High and Low tiers by the baseline model's accuracy, so the Low tier isolates precisely the languages where the model's native competence is weakest and translation should help most. On this independent distribution the gate remains the strongest cost-aware method by a wide margin. On MGSM Low it recovers 91\% of the free-tool gain (lifting reward from a 51.4 baseline to 83.2, against the free upper bound of 86.2) at 57\% tool use, while flat and OTC recover only 40\% and 29\% (65.4 and 61.6), barely moving the tool (14\% and 17\%) and leaving reward stranded near the baseline. Global MMLU Low shows the same shape: the gate recovers 80\% of the gain (40.3 to 61.5, free 66.7) at roughly half the tool cost, versus 6\% and 3\% for flat and OTC. On the High tiers, where the model is already strong, the gate correctly refrains, matching the baseline at a fraction of the free tool's cost. That the entire pattern, tier-adaptive tool use, near-free reward recovery, and the same ordering of the gate well ahead of both ungated penalties, reproduces on benchmarks built independently of us is strong evidence that the learned policy keys on real cross-lingual competence rather than on properties of our pipeline.

\begin{table*}[t]
\centering
\small
\begin{tblr}{
  width = \textwidth,
  colspec = {ll X[c] X[c] X[c] X[c] X[c] X[c] X[c] X[c] X[c] X[c]},
  hline{1} = {1.5pt, solid},
  hline{Z} = {1.5pt, solid},
  hline{2} = {3-12}{0.4pt, solid},
  hline{3} = {3-12}{0.4pt, solid},
  hline{4} = {0.8pt, solid},
  hline{6} = {0.4pt, solid},
  cell{1}{3} = {c=10}{c},
  cell{2}{3} = {c=2}{c},
  cell{2}{5} = {c=2}{c},
  cell{2}{7} = {c=2}{c},
  cell{2}{9} = {c=2}{c},
  cell{2}{11} = {c=2}{c},
  cell{4}{1} = {r=2}{m},
  cell{6}{1} = {r=2}{m},
  row{1-3} = {m, font=\bfseries},
  column{4,6,8,10,12} = {bg=gray!15},
}
  & & \textbf{External Benchmarks} & & & & & & & & & \\
  \textbf{Bench} & \textbf{Tier} & \textbf{Baseline} & & \textbf{+RL Free} & & \textbf{+RL Flat} & & \textbf{+RL OTC} & & \textbf{+RL Gate} & \\
  & & R\% & C\% & R\% & C\% & R\% & C\% & R\% & C\% & R\% & C\% \\
  MGSM & High & 88.6 & 1 & 88.9 & 92 & 87.6 & 1 & 88.2 & 2 & 88.6 & 21 \\
  & Low & 51.4 & 4 & 86.2 & 93 & 65.4 & 14 & 61.6 & 17 & 83.2 & 57 \\
  MMLU & High & 71.8 & 0 & 77.4 & 72 & 73.0 & 0 & 73.8 & 0 & 74.5 & 23 \\
  & Low & 40.3 & 2 & 66.7 & 65 & 41.9 & 1 & 41.2 & 2 & 61.5 & 51 \\
\end{tblr}
\caption{External benchmark results. R\% = reward, C\% = tool cost. Tiers split by baseline: MGSM Low = Swahili, Telugu; MMLU Low = Hausa, Igbo, Shona, Somali, Swahili, Yoruba.}
\label{tab:external}
\end{table*}


\section{Scaling to 8B}
\label{app:8b}

Our main experiments use Qwen3-4B. To check that the findings hold at larger scale, we replicate the free and gated tool-use runs on \textbf{Qwen3-8B} under an identical setup (same data, judge, and GSPO hyperparameters). The confidence gate uses a slightly gentler setting at this scale ($K{=}7$, $\lambda{=}0.01$), since the stronger base model clears the competence bar more easily. Table~\ref{tab:cost_focused_8b} reports the 8B results in a Baseline / +RL Free / +RL Gate format, with Free and Gate evaluated at matched training progress (the same base, free, and gated conditions as the 4B main table, Table~\ref{tab:cost_focused}).

\begin{table*}[t]
\centering
\small
\begin{tblr}{
  width = \textwidth,
  rowsep = 1pt,
  colspec = {ll X[c] X[c] X[c] X[c] X[c] X[c]},
  vline{5} = {dashed},
  hline{1} = {1.5pt, solid},
  hline{Z} = {1.5pt, solid},
  hline{2} = {3-8}{0.4pt, solid},
  hline{3} = {3-8}{0.4pt, solid},
  hline{4} = {0.8pt, solid},
  hline{8} = {0.4pt, solid},
  hline{11} = {0.4pt, solid},
  hline{14} = {0.4pt, solid},
  hline{16} = {0.4pt, solid},
  cell{1}{3} = {c=6}{c},
  cell{2}{3} = {c=2}{c},
  cell{2}{5} = {c=2}{c},
  cell{2}{7} = {c=2}{c},
  cell{4}{1} = {r=4}{m},
  cell{8}{1} = {r=3}{m},
  cell{11}{1} = {r=3}{m},
  cell{14}{1} = {r=2}{m},
  cell{16}{1} = {r=4}{m},
  row{1-3} = {m, font=\bfseries},
  column{4,6,8} = {bg=gray!15},
}
  & & \textbf{Tool} & & & & & \\
  \textbf{Dom} & \textbf{Tier} & \textbf{Baseline} & & \textbf{+RL Free} & & \textbf{+RL Gate} & \\
  & & R\% & C\% & R\% & C\% & R\% & C\% \\
  Math & High & 67.0 & 0.3 & 68.0 & 50.0 & 67.3 & 22.3 \\
  & Low & 31.5 & 1.6 & 66.5 & 81.1 & 65.0 & 61.8 \\
  & XLow & 46.0 & 0.8 & 63.0 & 76.5 & 63.0 & 53.8 \\
  & Synth & 16.0 & 17.5 & 58.0 & 87.5 & 56.0 & 74.5 \\
  QA & High & 83.3 & 0.0 & 85.7 & 58.5 & 85.3 & 9.8 \\
  & Low & 47.3 & 14.4 & 79.8 & 81.2 & 77.7 & 52.8 \\
  & XLow & 57.5 & 10.8 & 70.5 & 80.5 & 71.0 & 49.0 \\
  IF & High & 73.2 & 0.0 & 84.8 & 1.5 & 80.7 & 0.0 \\
  & Low & 49.2 & 1.2 & 64.9 & 13.2 & 64.9 & 0.9 \\
  & XLow & 54.5 & 0.2 & 69.0 & 7.8 & 68.7 & 0.5 \\
  Summ & High & 56.7 & 3.2 & 61.0 & 66.8 & 59.4 & 46.8 \\
  & Low & 6.2 & 29.5 & 41.9 & 97.2 & 39.1 & 97.2 \\
  Trans & High & 72.1 & 6.8 & 90.9 & 61.4 & 91.8 & 56.4 \\
  & Low & 14.0 & 13.0 & 68.2 & 57.5 & 70.0 & 58.6 \\
  & XLow & 19.3 & 15.2 & 56.1 & 58.3 & 55.4 & 63.9 \\
  & Synth & 25.5 & 72.5 & 68.0 & 100.0 & 67.2 & 99.0 \\
\end{tblr}
\caption{\textbf{8B tool results} (Qwen3-8B). R\% = reward, C\% = tool usage (max 2 calls; gray columns). Baseline = untrained model; +RL Free/Gate = after RL (matched training progress). Same setup as the 4B main table (Table~\ref{tab:cost_focused}); the 8B gate uses $K{=}7$, $\lambda{=}0.01$.}
\label{tab:cost_focused_8b}
\end{table*}

The 8B replication shows the same qualitative pattern as 4B. RL with the free tool lifts reward across all tiers, with the largest gains on low-resource and synthetic conditions where native competence is weakest. The gate then reproduces its central effect at scale: it cuts tool use sharply on Math and QA (for example, QA High cost drops from 58.5\% to 9.8\%, and Math High from 50.0\% to 22.3\%) while holding reward essentially flat (all reward changes within 2.9 points, and near zero on High and Low tiers). As at 4B, instruction following stays at near-zero tool use and translation remains high, and the two domains where reward dips slightly, IF High and Summarization Low, are the same weak spots seen at 4B. The gate thus preserves free-tool reward at substantially lower cost at 8B, confirming the mechanism is not specific to the smaller model.

\end{document}